%% file: main.tex
\documentclass[10pt,twocolumn,letterpaper]{article}

 \usepackage{cvpr}              
\usepackage{bm}

\usepackage{algorithm}
\usepackage{algorithmic}
\usepackage{multirow}
\usepackage{float, graphicx}
\usepackage{makecell}
\usepackage{cuted}
\usepackage{capt-of}
\usepackage{threeparttable}

\definecolor{cvprblue}{rgb}{0.21,0.49,0.74}
\usepackage[pagebackref,breaklinks,colorlinks,allcolors=cvprblue]{hyperref}

\def\paperID{4212} 
\def\confName{CVPR}
\def\confYear{2025}

\title{CoA: Towards Real Image Dehazing via Compression-and-Adaptation}

\author{Long Ma$^{1}$
	 \and Yuxin Feng$^{2}$ 
	 \and Yan Zhang$^{2}$ 
	  \and Jinyuan Liu$^{1}$
	 \and Weimin Wang$^{1}$
	 \and Guang-Yong Chen$^{3}$
	\and Chengpei Xu$^{4,}\thanks{Corresponding author.}$
	\and Zhuo Su$^{2}$
\and  \normalsize $^{1}$Dalian University of Technology$\;\;$
 $^2$Sun Yat-sen University$\;\;$
 $^3$Fuzhou University$\;\;$
 $^4$University of New South Wales\\
{\tt\small malone94319@gmail.com,\{fengyx26,zhangy2779\}@mail2.sysu.edu.cn,atlantis918@hotmail.com,}\\
{\tt\small Chengpei.Xu@unsw.edu.au,wangweimin@dlut.edu.cn,gychen@fzu.edu.cn,suzhuo3@mail.sysu.edu.cn}
}
\begin{document}
\maketitle
\input{sec/0_abstract}    
\input{sec/1_intro}
\input{sec/2_formatting}

{
    \small
    \bibliographystyle{ieeenat_fullname}
    \bibliography{main}
}


\end{document}

%% file: sec/0_abstract.tex
\thispagestyle{empty}
\begin{strip}
	\vspace{-1.6cm}
	\centering
	\begin{tabular}{c} 
		\includegraphics[width=0.98\textwidth]{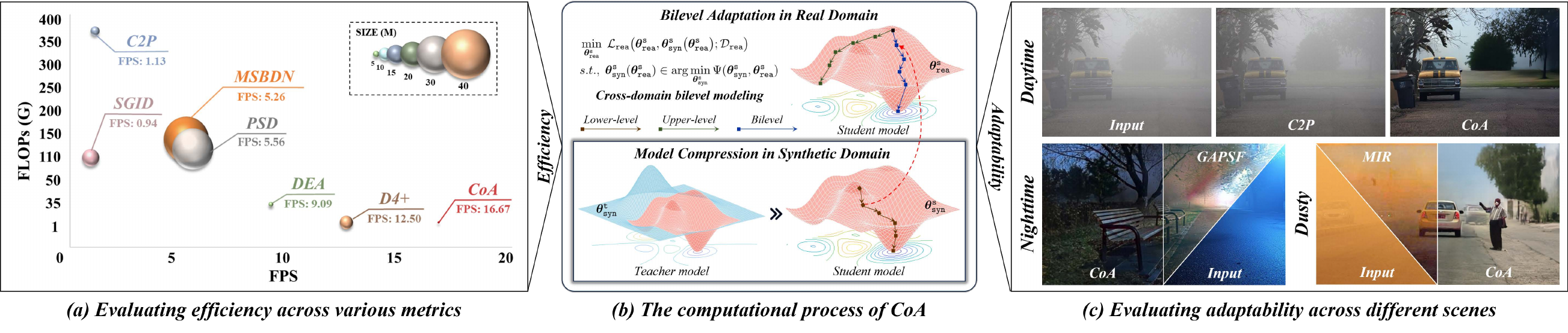}\\
	\end{tabular}
	\vspace{-0.4cm}
	\captionof{figure}{\textbf{Performance evaluation}. The proposed CoA incorporates model compression in synthetic domain for efficiency and bilevel adaptation in real domain for adaptability, as illustrated in the central sub-figure (b). The left sub-figure (a) presents efficiency across various metrics, clearly showing that our CoA outperforms others by a significant margin. The right sub-figure (c) shows adaptability across different scenes, where it is evident that our CoA consistently performs excellently. }
	\label{fig:FirstFigure}
	\end{strip}

\begin{abstract}

Learning-based image dehazing algorithms have shown remarkable success in synthetic domains. However, real image dehazing is still in suspense due to computational resource constraints and the diversity of real-world scenes. Therefore, there is an urgent need for an algorithm that excels in both efficiency and adaptability to address real image dehazing effectively. 
This work proposes a \textbf{Co}mpression-and-\textbf{A}daptation (\textbf{CoA}) computational flow to tackle these challenges from a divide-and-conquer perspective. First, model compression is performed in the synthetic domain to develop a compact dehazing parameter space, satisfying efficiency demands. Then, a bilevel adaptation in the real domain is introduced to be fearless in unknown real environments by aggregating the synthetic dehazing capabilities during the learning process. 
Leveraging a succinct design free from additional constraints, our CoA exhibits domain-irrelevant stability and model-agnostic flexibility,  effectively bridging the model chasm between synthetic and real domains to further improve its practical utility. Extensive evaluations and analyses underscore the approach's superiority and effectiveness. The code is publicly available at~\url{https://github.com/fyxnl/COA}.

\end{abstract}

%% file: sec/1_intro.tex
\section{Introduction}
\label{sec:intro}
Image dehazing tasks aim to utilize the atmospheric optical model and advanced computing theory to achieve efficient restoration of scene information and image details through inverse imaging solution or data-driven model learning. It has drawn much attention in multiple emerging computer vision areas recently~\cite{DBLP:conf/cvpr/ZhengZHD023}~\cite{DBLP:journals/tip/LingCTJC23}~\cite{DBLP:journals/pami/LiuLSZ23}.
Similar to other low-level vision tasks~\cite{ma2022toward}~\cite{DBLP:journals/pami/FuXZLWZ23}, image dehazing technology has evolved from an early stage focused on improving synthetic data metrics to a direction aimed at effectively generalizing across various real-world haze scenarios. 
Current approaches face dual challenges: efficiency-oriented methods achieve real-time processing but lack scene adaptability, while adaptability-focused techniques dynamically adjust strategies at the cost of high computational complexity that hinders deployment in resource-constrained environments. This paper addresses these limitations through novel architectural innovations in both computational efficiency and adaptive scene understanding.

\subsection{Related Works}

\textbf{Efficiency-oriented Dehazing Methods.}
This work focuses on enhancing computational efficiency through optimization algorithms, model simplification, or machine learning methods to enable faster processing under limited resources. Broadly, dehazing methods can be divided into image enhancement-based, physics-based, and image layering-based approaches. Image enhancement-based dehazing methods~\cite{DBLP:journals/tip/LiTTWW21} treat the task as an image improvement problem, increasing contrast and color accuracy with techniques like histogram equalization and color correction. Physics-based methods~\cite{he2010single},~\cite{DBLP:journals/pami/LiuLSZ23} are more specific, leveraging hazy image priors and constraints to aid inverse problem solving. Image layering-based approaches, such as that of Zhang \textit{et al}.~\cite{zhang2018densely}, increase efficiency by breaking down the image into scales or layers, progressively refining transmittance maps and enhancing detail using multilevel pyramid structures and dense connectivity. Despite their computational and real-time advantages, they struggle with stability and consistent quality, especially in complex real-world haze conditions, often resulting in partial dehazing and insufficient color clarity.

\noindent \textbf{Adaptability-focused Dehazing Methods.}
Adaptive dehazing methods tailor their strategies based on specific image attributes to improve dehazing effectiveness. These methods encompass three main types: local feature-based, atmospheric scattering model-based, and domain adaptation-based approaches. Local feature-based methods~\cite{DBLP:journals/tmm/GuiCHTK24} adaptively adjust dehazing intensity according to local contrast and brightness variations, which helps restore image details effectively.
Atmospheric scattering model-based adaptive dehazing methods~\cite{DBLP:journals/tip/JuDGRT21} dynamically adjust the transmittance estimation by analyzing different regions of the image to achieve more accurate restoration. These methods can, to some extent, balance both efficiency and adaptability. Domain adaptation-based dehazing methods~\cite{DBLP:conf/cvpr/WuDGCL23,DBLP:conf/ijcai/LiangWZ0R22} have gradually emerged in the field of image dehazing in recent years. These methods learn a shared feature representation, enabling the model to simultaneously adapt to the features of both the source domain and the target domain. 
Although these methods yield superior dehazing results, their complex models and dynamic image-specific adjustments often increase computational costs, potentially impacting real-time applicability.

\renewcommand{\algorithmicrequire}{ \textbf{Require:}} 
\renewcommand{\algorithmicensure}{ \textbf{Ensure:}} 

\subsection{Our Contributions}
To address the challenges of efficiency and adaptability in real image dehazing, this work proposes a new \textbf{Co}mpression-and-\textbf{A}daptation (\textbf{CoA}) scheme as shown in Fig.~\ref{fig:FirstFigure} (b). The predefined model is compressed in the synthetic domain to ensure computational efficiency (refer to Fig.~\ref{fig:FirstFigure} (a)). Further, a bilevel adaptive learning is developed to leverage the dehazing capability acquired in the synthetic domain to improve adaptability (see Fig.~\ref{fig:FirstFigure} (c)) in the real domain. Our key contributions are summarized as 
\begin{itemize}
	\item Following the divide-and-conquer paradigm, we propose a new CoA learning strategy that first compresses the predefined model and then adapts it to unseen scenes, offering an efficient solution for real image dehazing. To the best of our knowledge, this is the first work to successfully integrate both efficiency and adaptability, achieving the best of both worlds in real image dehazing.
	\item To adapt to unlabeled, diverse real-world scenes, we propose a new cross-domain bilevel model designed to learn dehazing parameters for the real domain constrained by synthetic dehazing learning. Additionally, we derive a bilevel adaptive learning scheme that effectively harnesses the synthetic dehazing capacity, preventing instability and enhancing adaptability.
	\item Benefiting from succinct design free from additional constraints, CoA shows two key properties on three standard real benchmarks. It shows stability across synthetic domains, improving 14.2\% on average across four metrics (evaluated on three domains). CoA also offers flexibility with dehazing models, reducing computational cost by at least 73\% in parameters (tested on three methods).
\end{itemize}

%% file: sec/2_formatting.tex
\section{The Proposed Method}
In this section, we first rethinking real image dehazing, highlighting two key practical needs. We then propose model compression and bilevel adaptation to meet these needs, followed by an overview of the network architecture and training loss.

\begin{figure*}[t]
	\centering
	\footnotesize
	\begin{tabular}{c} 
		\includegraphics[width=0.98\textwidth]{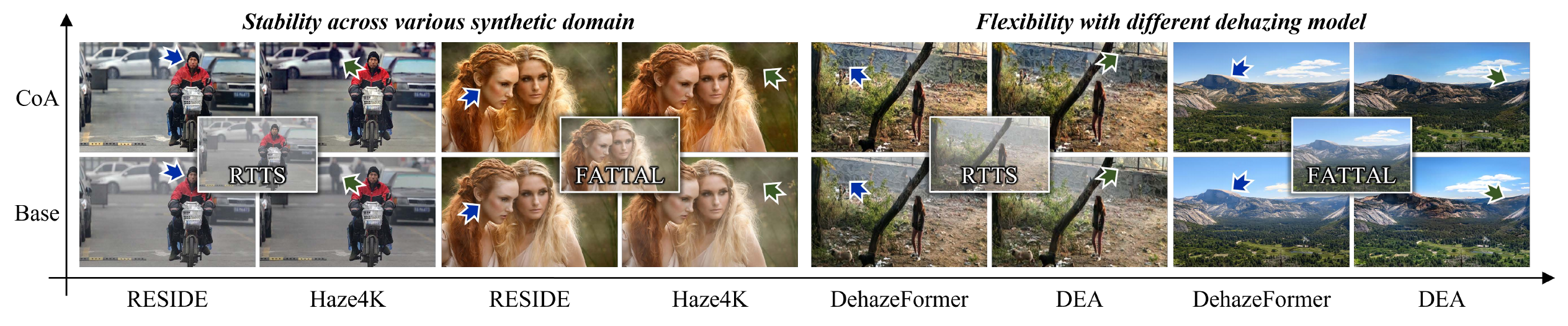}\\
	\end{tabular}
    \vspace{-0.3cm}
	\caption{\textbf{Qualitative comparisons of algorithmic properties}. Arrows indicate specific regions that highlight visible differences.}
	\label{fig:PropertyQual}
\end{figure*}

\subsection{Rethinking Real Image Dehazing}\label{sec:rethinking}
Despite the strong performance of advanced image dehazing techniques~\cite{wang2024odcr,zhang2024depth,liu2024sdcnet} on synthetic datasets~\cite{li2018benchmarking,liu2021synthetic,feng2024bridging}, these datasets often rely on simplifying assumptions, resulting in a substantial gap between synthetic and real-world domains. Addressing the core challenges in real-world image dehazing involves two key aspects:
\begin{itemize}
		\item \textbf{\textit{Efficiency}}: \textit{Edge devices often have limited computational resources compared to systems with ample capacity, requiring dehazing models to be highly efficient.} 
		\item \textbf{\textit{Adaptability}}: \textit{Real-world imaging conditions vary dynamically. Dehazing models must therefore be adaptable to maintain reliable performance.} 
\end{itemize}

The latter is more challenging, relying on effective learning mechanisms like unsupervised learning~\cite{golts2019unsupervised}. However, current methods struggle with unsupervised learning due to difficulty in capturing statistical regularities, making transferring dehazing capabilities from the synthetic to real domain a viable alternative.
To provide an intuitive presentation of the goal upon the above demands, we build
\begin{equation}\label{eq:rethink}
	\begin{aligned}
		&\min_{\bm{\theta}_{\mathtt{real}}} f\Big(\bm{\theta}_{\mathtt{real}}\big(\bm{\theta}_{\mathtt{syn}}\big)\Big),\\
		&s.t., \left\{
		\begin{aligned}
			&\kappa\big(\bm{\theta}_{\mathtt{real}}\big)<\kappa\big(\bm{\theta}_{\mathtt{syn}}\big), (\mathtt{efficiency})\\
			&\zeta\big(\bm{\theta}_{\mathtt{real}}\big)>\zeta\big(\bm{\theta}_{\mathtt{syn}}\big), (\mathtt{adaptability})\\
		\end{aligned}
		\right.
	\end{aligned}
\end{equation}
where $\bm{\theta}_{\mathtt{syn}}$ and $\bm{\theta}_{\mathtt{real}}$ represent parameters in synthetic and real domain, respectively. The function $\kappa(\cdot)$ and $\zeta(\cdot)$ denote the evaluator for efficiency and adaptability, respectively. 

To achieve this, we propose a divide-and-conquer strategy: the first phase focuses on model compression for efficiency, while the second phase employs bilevel adaptation for adaptability. Details are provided below.
\subsection{\underline{MoC}:Model Compression in Synthetic Domain}
The main challenge in model compression is transferring dehazing capabilities from a large-scale parameter space to a smaller one. Here we propose a composite loss function\footnote{Details of the specific loss functions can be found in Sec.~\ref{sec:architecture-loss}.}  with context alignment to realize it.
\begin{equation}\label{eq:MoC}
	\min_{\bm{\theta}_{\mathtt{syn}}^{\mathtt{s}}}\mathcal{L}_{\mathtt{syn}}(\bm{\theta}_{\mathtt{syn}}^{\mathtt{s}})+\mathcal{L}_{\mathtt{a}}\big(\bm{\theta}_{\mathtt{syn}}^{\mathtt{s}},\bm{\theta}_{\mathtt{syn}}^{\mathtt{t}}\big),
\end{equation}
where $\bm{\theta}_{\mathtt{syn}}^{\mathtt{t}}$ and $\bm{\theta}_{\mathtt{syn}}^{\mathtt{s}}$ are the pre-trained teacher model and desired student model in the synthetic domain, respectively. The function $\mathcal{L}_{\mathtt{syn}}$ and $\mathcal{L}_{\mathtt{a}}$ represent a group of supervised losses and the context alignment loss, respectively.

During model compression, the student model follows the teacher model's training strategy on paired synthetic data. To maximize the retention of the teacher model's capabilities, we introduce a context alignment loss ($\mathcal{L}_{\mathtt{a}}$) that aligns the outputs of the teacher and student encoders at corresponding positions. By weighting the loss functions across layers, this approach enables fine-grained feature alignment, improving knowledge transfer and model optimization. The layered alignment ensures both global feature distributions and layer-specific details are accurately matched, resulting in more precise compression of the teacher model into the student model.

\begin{table*}[t]
	\renewcommand\arraystretch{1.2} 
	\setlength{\tabcolsep}{1.7mm}
	\centering
	\footnotesize
	\begin{tabular}{|l|l||cc|cc|cc||cc|cc|cc|}
		\hline
		\multirow{3}{*}{\makecell{Testing\\Benchmark}}&\multirow{3}{*}{Metrics}&\multicolumn{6}{c||}{\textbf{\textit{Stability across various synthetic domain}}}&\multicolumn{6}{c|}{\textbf{\textbf{\textit{Flexibility with different dehazing model}}}}\\
		\cline{3-14}
		~&~&\multicolumn{2}{c|}{RESIDE}&\multicolumn{2}{c|}{Haze4K}&\multicolumn{2}{c||}{THaze}&\multicolumn{2}{c|}{MSBDN}&\multicolumn{2}{c|}{DehazeFormer}&\multicolumn{2}{c|}{DEA}\\
		\cline{3-14}
		~&~&Base&CoA&Base&CoA&Base&CoA&Base&CoA&Base&CoA&Base&CoA\\
		\hline
		\multirow{4}{*}{RTTS~\cite{DBLP:journals/tip/LiRFTFZW19}}&FADE$\downarrow$&1.4753&\textbf{1.2687}&1.7368&\textbf{1.0850}&1.2078&\textbf{0.8594}&1.3354&\textbf{1.1556}&1.1866&\textbf{0.9325}&1.2117&\textbf{0.9332}\\
		&PM2.5$\downarrow$ &163.77&\textbf{139.04}&170.03&\textbf{110.84}&131.77&\textbf{88.871}&135.68&\textbf{95.793}&126.77&\textbf{115.00}&121.98&\textbf{89.813}\\
		&Entropy$\uparrow$&7.2963&\textbf{7.3458}&7.3446&\textbf{7.4637}&7.4319&\textbf{7.5798}&7.3968&\textbf{7.5382}&7.4815&\textbf{7.5511}&7.4623&\textbf{7.5426}\\
		&BIQME$\uparrow$&0.5562&\textbf{0.5639}&0.5511&\textbf{0.5748}&0.5773&\textbf{0.5932}&0.5696&\textbf{0.5819}&0.5733&\textbf{0.5928}&0.5789&\textbf{0.5860}\\
		\hline
		\multirow{4}{*}{URHI~\cite{DBLP:journals/tip/LiRFTFZW19}}&FADE$\downarrow$&1.4463&\textbf{1.3013}&1.7263&\textbf{1.0278}&1.3146&\textbf{0.9272}&1.3488&\textbf{0.9549}&1.3263&\textbf{1.1726}&1.2796&\textbf{1.0321}\\
		&PM2.5$\downarrow$ &164.36&\textbf{145.01}&172.98&\textbf{119.97}&140.35&\textbf{98.793}&146.43&\textbf{99.856}&141.11&\textbf{125.87}&133.97&\textbf{108.94}\\
		&Entropy$\uparrow$&7.2867&\textbf{7.3152}&7.3711&\textbf{7.4657}&7.4652&\textbf{7.5928}&7.4144&\textbf{7.5574}&7.5009&\textbf{7.5817}&7.4920&\textbf{7.5658}\\
		&BIQME$\uparrow$&0.5499&\textbf{0.5542}&0.5481&\textbf{0.5681}&0.5821&\textbf{0.5961}&0.5727&\textbf{0.5832}&0.5757&\textbf{0.5948}&0.5786&\textbf{0.5863}\\
		\hline
		\multirow{4}{*}{FATTAL~\cite{fattal2014dehazing}}&FADE$\downarrow$&0.5136&\textbf{0.4778}&0.5828&\textbf{0.3734}&0.4203&\textbf{0.3142}&0.4315&\textbf{0.3327}&0.4271&\textbf{0.3875}&0.4270&\textbf{0.3976}\\
		&PM2.5$\downarrow$ &81.738&\textbf{80.289}&81.401&\textbf{110.97}&62.312&\textbf{91.639}&69.272&\textbf{104.56}&78.093&\textbf{60.693}&81.387&\textbf{80.723}\\
		&Entropy$\uparrow$&7.3847&\textbf{7.4575}&7.3999&\textbf{7.5006}&7.4482&\textbf{7.5858}&7.4121&\textbf{7.5460}&7.4566&\textbf{7.5610}&7.4315&\textbf{7.5569}\\
		&BIQME$\uparrow$&0.5645&\textbf{0.5802}&0.5554&\textbf{0.5919}&0.5945&\textbf{0.6189}&0.5807&\textbf{0.6055}&0.5927&\textbf{0.6117}&0.5903&\textbf{0.6097}\\
		\hline
	\end{tabular}
    \vspace{-0.2cm}
	\caption{\textbf{Quantitative comparison of algorithmic properties}. In the left portion, our designed architecture serves as the baseline teacher model. In the right portion, the THaze dataset is used as the synthetic domain.}\label{tab:PropertyQuan}
        \vspace{-0.4cm}
\end{table*}

\begin{algorithm}[t]
	\caption{Learning via Compression-and-Adaptation}\label{alg:BAL}
	\begin{algorithmic}[1]
		\REQUIRE The pre-trained teacher model $\bm{\theta}_{\mathtt{syn}}^{\mathtt{t}}$, the initial parameter $\bm{\theta}_{\mathtt{syn}}^{\mathtt{s}}$, step-sizes $\eta_{\mathtt{syn}}^m, \eta_{\mathtt{syn}}^b$, coefficient $\alpha$, iteration numbers $N$ in MoC phase and $T$ in BiA phase.
		\ENSURE The optimal parameters $\bm{\theta}_{\mathtt{rea}}^{\mathtt{s}}$.
		\STATE \% The MoC Phase
		\FOR{$n=0:N-1$}
		\STATE Calculate the gradient $\bm{g}$ of Eq.~\eqref{eq:MoC}.
		\STATE $\bm{\theta}_{\mathtt{syn}}^{\mathtt{s}}(n+1)=\bm{\theta}_{\mathtt{syn}}^{\mathtt{s}}(n)-\eta_{\mathtt{syn}}^m\bm{g}(\bm{\theta}_{\mathtt{syn}}^{\mathtt{s}}(n),\bm{\theta}_{\mathtt{syn}}^{\mathtt{t}})$.
		\ENDFOR
		\STATE Initialize the dehazing model's parameter in the real domain by $\bm{\theta}_{\mathtt{rea}}^{\mathtt{s}}=\bm{\theta}_{\mathtt{syn}}^{\mathtt{s}}(N)$.
		\STATE \% The BiA Phase
		\FOR{$t=0:T-1$}
		\STATE \% Update the lower-level parameter $\bm{\theta}_{\mathtt{syn}}^{\mathtt{s}}$ in $\mathcal{D}_{\mathtt{rea}}$.
		\STATE $\bm{\theta}_{\mathtt{syn}}^{\mathtt{s}}(t+1)=\bm{\theta}_{\mathtt{syn}}^{\mathtt{s}}(t)-\eta_{\mathtt{syn}}^b\frac{\partial(\mathcal{L}_{\mathtt{rea}}+\mathcal{L}_{\mathtt{1}})}{\partial\bm{\theta}_{\mathtt{syn}}^{\mathtt{s}}}$.
		\STATE \% Update the upper-level parameter $\bm{\theta}_{\mathtt{rea}}^{\mathtt{s}}$ by EMA.
		\STATE $\bm{\theta}_{\mathtt{rea}}^{\mathtt{s}}(t+1)=\alpha\bm{\theta}_{\mathtt{rea}}^{\mathtt{s}}(t)+(1-\alpha)\bm{\theta}_{\mathtt{syn}}^{\mathtt{s}}(t+1)$.
		\ENDFOR
		\RETURN $\bm{\theta}_{\mathtt{rea}}^{\mathtt{s}}(T)$.
	\end{algorithmic}
\end{algorithm}
\subsection{\underline{BiA}:Bilevel Adaptation to Real Domain}
After compressing the model for efficiency, we address adaptation to real-world scenes by proposing a cross-domain bilevel model. This is followed by a bilevel adaptive learning scheme to enable smooth transition from the synthetic to real domain.

\subsubsection{Cross-Domain Bilevel Modeling}
The gap between the synthetic and real domains makes adaptation challenging. The dehazing parameters in the synthetic and real domains ($\bm{\theta}_{\mathtt{rea}}^{\mathtt{s}}$) have a nested subordinate relationship. This coupling can be effectively modeled using bilevel programming techniques~\cite{franceschi2018bilevel,liu2021investigating}. From the perspective of hyperparameter optimization, 
our cross-domain bilevel model is written as:
\begin{equation}\label{eq:bilevel}
	\begin{aligned}
    &\min_{\bm{\theta}_{\mathtt{rea}}^{\mathtt{s}}}\;\mathcal{L}_{\mathtt{rea}}\big(\bm{\theta}_{\mathtt{rea}}^{\mathtt{s}},\bm{\theta}_{\mathtt{syn}}^{\mathtt{s}}\big(\bm{\theta}_{\mathtt{rea}}^{\mathtt{s}}\big);\mathcal{D}_{\mathtt{rea}}\big),\\
		&s.t.,\;\bm{\theta}_{\mathtt{syn}}^{\mathtt{s}}(\bm{\theta}_{\mathtt{rea}}^{\mathtt{s}})\in\arg\min_{\bm{\theta}_{\mathtt{syn}}^{\mathtt{s}}}\Psi(\bm{\theta}_{\mathtt{rea}}^{\mathtt{s}},\bm{\theta}_{\mathtt{syn}}^{\mathtt{s}}),\\ 
	\end{aligned}
\end{equation}
where 
\begin{equation}
\Psi(\bm{\theta}_{\mathtt{rea}}^{\mathtt{s}},\bm{\theta}_{\mathtt{syn}}^{\mathtt{s}})=\mathcal{L}_{\mathtt{rea}}(\bm{\theta}_{\mathtt{syn}}^{\mathtt{s}};\mathcal{D}_{\mathtt{rea}})+\mathcal{L}_{\mathtt{1}}(\bm{\theta}_{\mathtt{syn}}^{\mathtt{s}},\bm{\theta}_{\mathtt{rea}}^{\mathtt{s}}),
\end{equation}
where $\mathcal{L}_{\mathtt{rea}}$ and $\mathcal{L}_{\mathtt{1}}$ represent the loss constraints\footnote{Their specific forms can be found in Sec.~\ref{sec:architecture-loss}.} for real domain and alignment. The datasets $\mathcal{D}_{\mathtt{rea}}$ and $\mathcal{D}_{\mathtt{syn}}$ are captured in the real and synthetic domains, respectively.

Actually, suppose an effective loss $\mathcal{L}_{\mathtt{rea}}$ could be designed to constrain the output in the real domain, the adaptation might be achieved solely through the upper-level model (initialized with $\bm{\theta}_{\mathtt{syn}}^{\mathtt{s}}$). However, as discussed in~Sec.~\ref{sec:rethinking}, the challenge of accurately modeling real haze distributions makes this process difficult to implement. To this end, we introduce lower-level sub-optimization to ensure a feasible solution and maintain the correct optimization trajectory. Experimental validation can be seen in Sec.~\ref{sec:necessity}.

\begin{figure*}[t]
	\centering
	\begin{tabular}{cc} 
		\includegraphics[width=0.47\textwidth]{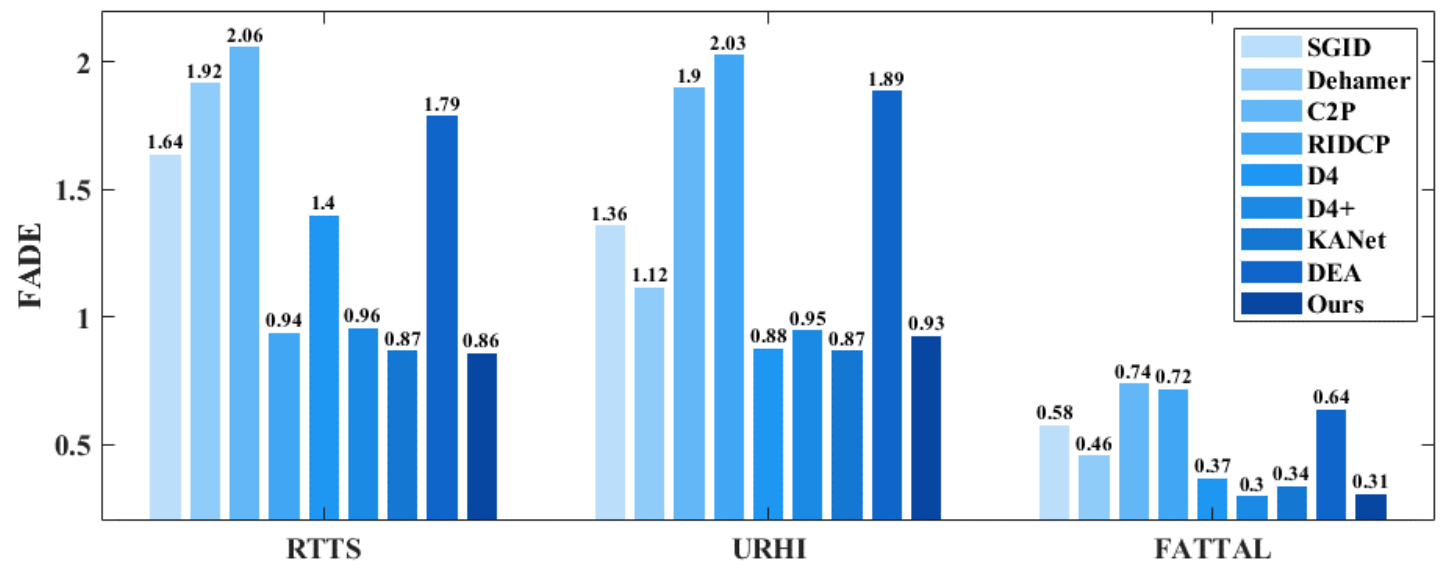}&
		\includegraphics[width=0.47\textwidth]{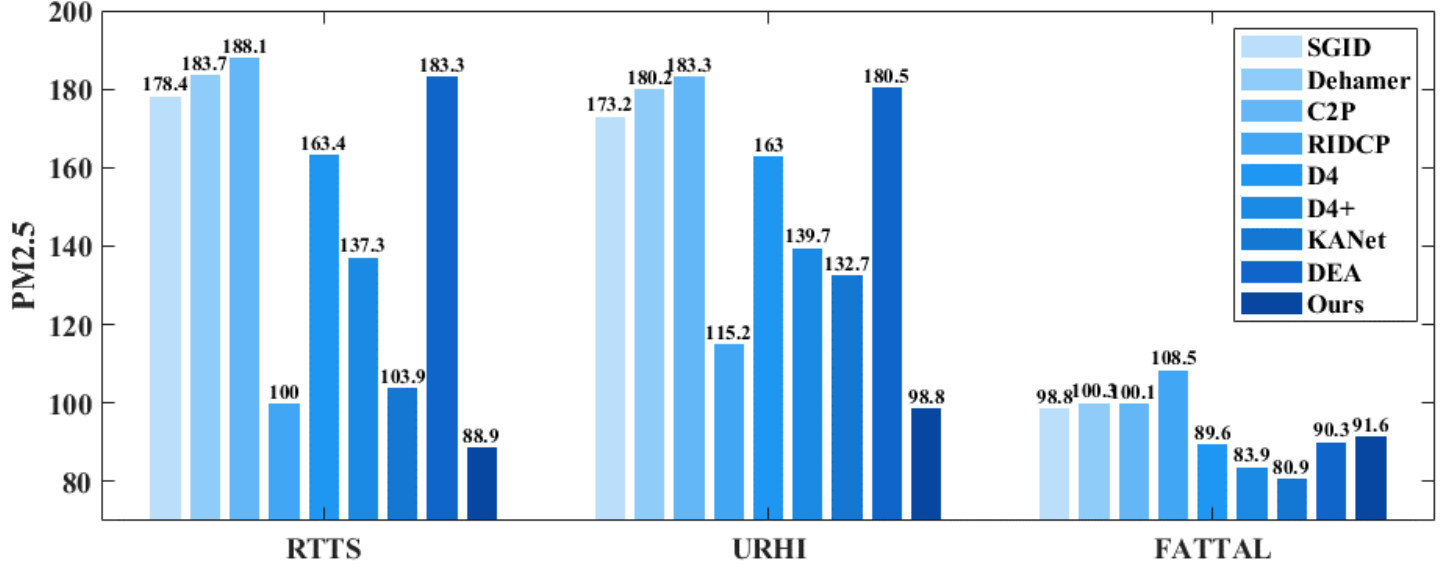}\\
		\includegraphics[width=0.47\textwidth]{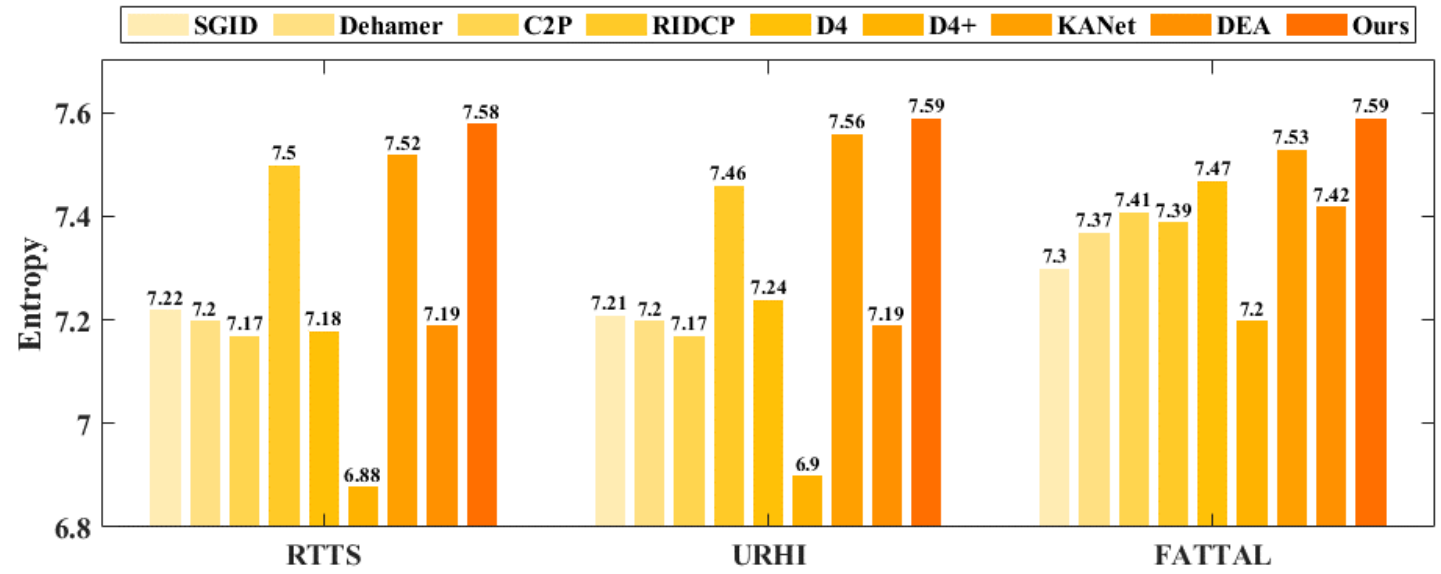}&
		\includegraphics[width=0.47\textwidth]{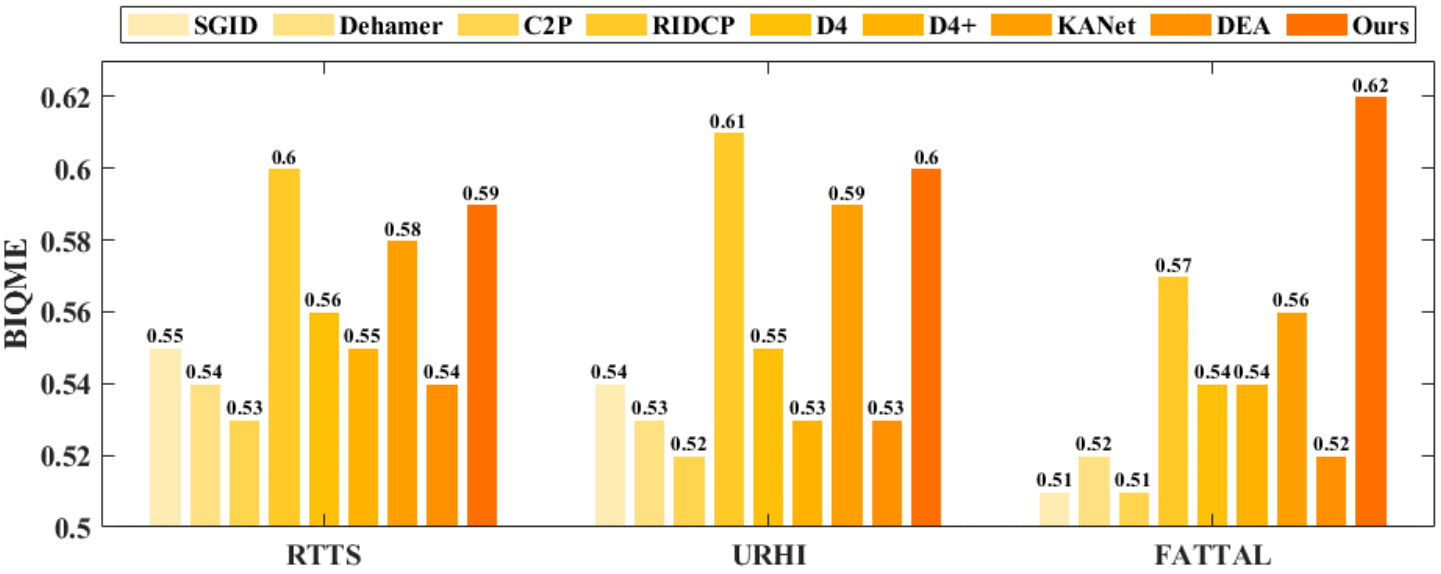}\\
	\end{tabular}
        \vspace{-0.2cm}
	\caption{\textbf{Quantitative comparison on three real-world datasets.} Four no-reference image quality assessments were calculated.}
	\label{fig:Quantitative}
        \vspace{-0.2cm}
\end{figure*}
\begin{figure*}[t]
	\centering
	\begin{tabular}{c} 
		\includegraphics[width=0.98\textwidth]{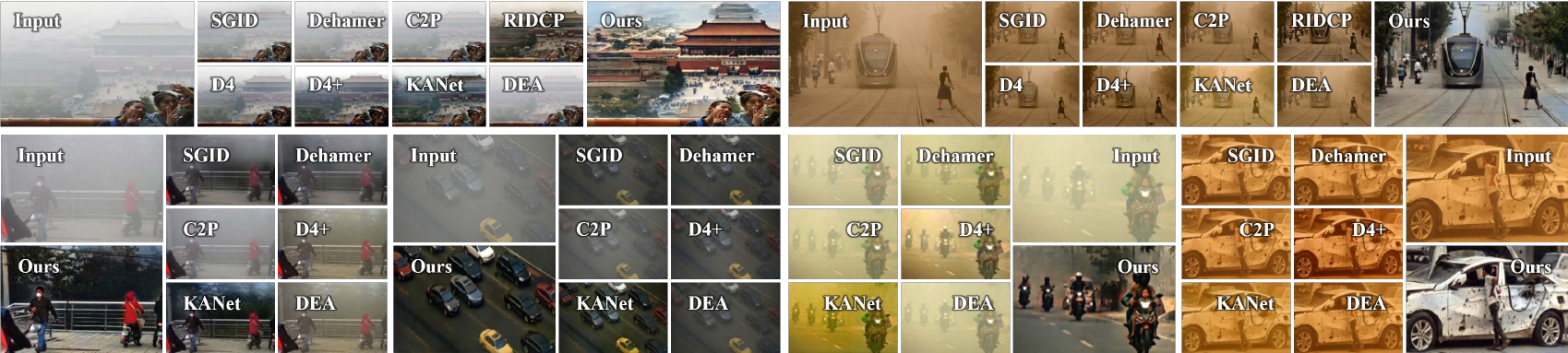}\\
	\end{tabular}
        \vspace{-0.2cm}
	\caption{\textbf{Qualitative comparisons on \textbf{\textit{daytime haze and dusty}} scenes}. All these observations come from RTTS and URHI datasets.}
	\label{fig:Daytime}
        \vspace{-0.3cm}
\end{figure*}

\subsubsection{Bilevel Adaptive Learning}
Without compromising the model's dehazing capability in the synthetic domain, narrowing the performance gap between the synthetic and real domains is a central focus. Inspired by the Exponential Moving Average (EMA) method~\cite{cai2021exponential}, we draw on its smoothing characteristics during the training process and introduce a dynamic weight adjustment mechanism to achieve a balance in learning between the synthetic and real domains, formulated as
\begin{equation}
	\left\{
	\begin{aligned}
		&\bm{\theta}_{\mathtt{syn}}^{\mathtt{s}}(t+1)=\bm{\theta}_{\mathtt{syn}}^{\mathtt{s}}(t)-\eta_{\mathtt{syn}}^b\frac{\partial(\mathcal{L}_{\mathtt{rea}}+\mathcal{L}_{\mathtt{1}})}{\partial\bm{\theta}_{\mathtt{syn}}^{\mathtt{s}}},\\
		&\bm{\theta}_{\mathtt{rea}}^{\mathtt{s}}(t+1)=\alpha\bm{\theta}_{\mathtt{rea}}^{\mathtt{s}}(t)+(1-\alpha)\bm{\theta}_{\mathtt{syn}}^{\mathtt{s}}(t+1),\\
	\end{aligned}
	\right.
\end{equation}
where $\alpha$ represents the moving average coefficient, and $\mathcal{L}_{\mathtt{1}}$ denotes the $\ell_1$-norm supervised loss used to reduce the risk of the model overfitting to the real domain. Specifically, during each update, $\bm{\theta}_{\mathtt{rea}}^{\mathtt{s}}$ is adjusted by $\bm{\theta}_{\mathtt{syn}}^{\mathtt{s}}$, allowing the model to strike a balance between the synthetic and real domains throughout the learning process. The overall algorithm can be seen in Alg.~\ref{alg:BAL}. 
\subsection{Network Architecture and Training Loss}\label{sec:architecture-loss}
This section introduces the details about network architecture and training loss adopted in this work. 
\subsubsection{Network Architecture}
We emphasize that our primary focus is on developing an effective learning mechanism to enhance the model's adaptability to the real domain. Notably, our approach offers flexibility\footnote{For further details, please refer to Sec.~\ref{sec:property}.} with different dehazing models. To this end, we adopt a hybrid architectural design, incorporating a pre-trained Res2Net encoder~\cite{DBLP:conf/cvpr/FuLYCW21} with the decoder of multi-scale boosted dehazing network~\cite{dong2020multi}.
\subsubsection{Loss Functions in MoC Phase}
In the MoC process, to enable the student model to learn the features extracted by the teacher model more effectively and to enhance the quality the generated images. We define
\begin{equation}
	\mathcal{L}_{\mathtt{syn}}=\lambda _{su}\ell_{su}+\lambda _{ss}\ell_{ss}+\lambda _{pe}\ell_{pe},
\end{equation}
where $\ell_{su}, \ell_{ss}, \ell_{pe}$ denote $\ell_1$-norm supervised loss, SSIM loss, and perceptual loss, respectively. The coefficient $\lambda _{su}, \lambda _{ss}, \lambda _{pe}$ are positive balancing parameters. 

To more effectively enhance consistency between the student and teacher models in the feature space, the design of $\mathcal{L}_{\mathtt{a}}$ incorporates a similarity measure of feature distributions. Its formulation can be written as follows:
\begin{equation}
	\mathcal{L}_{\mathtt{a}}=\sum_{i=0}^{L-1}{w}_i \cdot \big( ( T_i-S_i ) ^2\big) ,
\end{equation}
where $w_i$ represents the weights of $i$-th layer in the teacher feature map $T_i$ and the student feature map $S_i$.

\subsubsection{Loss Functions in BiA Phase}
Here we incorporate the CLIP model to construct the training loss $\mathcal{L}_{\mathtt{rea}}$ to constrain the output in the real domain.
We select haze images $I_{\mathcal{H}}$ and clear images $I_{\mathcal{C}}$ as references, initialize corresponding prompts $T_{\mathcal{H}}$ and $T_{\mathcal{C}}$, and input them into the CLIP text encoder, while feeding the images into the image encoder. The sample categories are predicted using the text-image similarity in the CLIP space. To minimize the classification error between positive and negative samples, we employ the binary cross entropy loss to optimize the text prompt pair. After training the text prompts to effectively distinguish between real haze and clear images, we can derive $\mathcal{L}_{\mathtt{rea}}$ based on the contrastive similarity loss:
\begin{equation}
	\mathcal{L}_{\mathtt{rea}}=\frac{e^{cos\big(\Phi_{image}(I_\mathcal{R}),\Phi_{text}(T_\mathcal{H})\big)}}{\sum_{i\in\{\mathcal{H},\mathcal{C}\}}e^{cos\big(\Phi_{image}(I_\mathcal{R}),\Phi_{text}(T_i)\big)}},
\end{equation}
where $I_\mathcal{R}$ corresponds to the real haze image processed by the model obtained through the model compression process.
\begin{table*}[t]
	\renewcommand\arraystretch{1.2} 
	\setlength{\tabcolsep}{1.5mm}
	\centering
	\footnotesize
	\begin{tabular}{|l|c|c|c|c|c|c|c|c|c|c|c|c|}
		\hline
		\multirow{2}{*}{Method} & \multicolumn{4}{c|}{RTTS}&\multicolumn{4}{c|}{URHI}&\multicolumn{4}{c|}{FATTAL}\\
		\cline{2-13}
		~ & FADE$\downarrow$&PM2.5$\downarrow$ &Entropy$\uparrow$&BIQME$\uparrow$&FADE$\downarrow$&PM2.5$\downarrow$ &Entropy$\uparrow$&BIQME$\uparrow$&FADE$\downarrow$&PM2.5$\downarrow$ &Entropy$\uparrow$&BIQME$\uparrow$\\
		\hline
		{AirNet} &1.3342&155.84&7.2959&0.5471&1.6591&143.02&6.9143&0.5258&0.4063&76.114&7.3545&0.5315\\
		{WeatherDiff} &2.4102 &193.27&7.1443&0.5155&2.1429&180.26&7.1419&0.5102&0.8749&118.79&7.3621&0.4937\\
		{DiffUIR} &2.1305&189.61&7.1722&0.5303&1.9879&180.56&7.1811&0.5205&0.8405&116.68&7.3609&0.5049\\
		\hline
		{Ours} &{\textbf{0.8594}}&{\textbf{88.871}}&{\textbf{7.5798}}&{\textbf{0.5932}}&{\textbf{0.9272}}&{\textbf{98.793}}&{\textbf{7.5928}}&{\textbf{0.5961}}&{\textbf{0.3142}}&{\textbf{91.639}}&{\textbf{7.5858}}&{\textbf{0.6189}}\\
		\hline
	\end{tabular}
    \vspace{-0.2cm}
    \caption{\textbf{Quantitative comparison on \textbf{\textit{daytime haze}} scenes}. All compared methods are designed for multi-weather image restoration. }
    \label{tab:MultiTask}
\end{table*}

\begin{figure*}[t]
	\centering
	\begin{tabular}{c} 
		\includegraphics[width=0.98\textwidth]{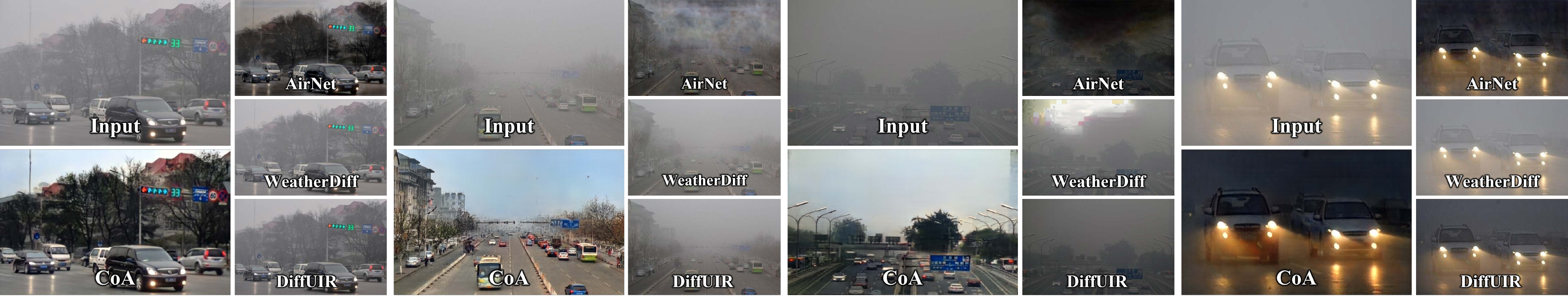}\\
	\end{tabular}
        \vspace{-0.2cm}
	\caption{\textbf{Qualitative comparisons corresponding to Table~\ref{tab:MultiTask}}. All examples are sourced from RTTS and URHI datasets.}
	\label{fig:MultiTask}
        \vspace{-0.3cm}
\end{figure*}

\begin{figure}[t]
	\centering
	\footnotesize
	\begin{tabular}{c} 
		\includegraphics[width=0.455\textwidth]{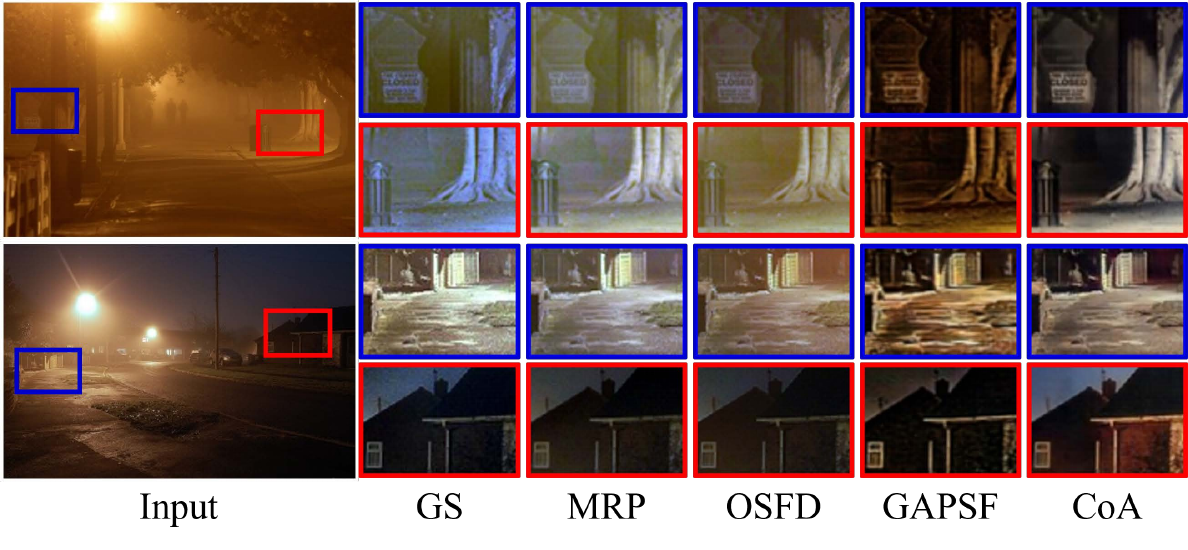}\\
		(a) Qualitative comparison 
	\end{tabular}
	\renewcommand\arraystretch{1.1} 
	\setlength{\tabcolsep}{0.8mm}
	\footnotesize
	\centering
	\begin{tabular}{|l|c|c|c|c|c|}
		\hline
		\multirow{2}{*}{Metrics}&\multirow{2}{*}{\makecell{GS\\\scriptsize{(\textit{ICCV'15})}}}&\multirow{2}{*}{\makecell{MRP\\\scriptsize{(\textit{CVPR'17})}}}&\multirow{2}{*}{\makecell{OSFD\\\scriptsize{(\textit{ACM MM'20})}}}&\multirow{2}{*}{\makecell{GAPSF\\\scriptsize{(\textit{ACM MM'23})}}}&\multirow{2}{*}{CoA}\\
		&&&&&\\
		\hline
		PM2.5↓ &{\underline{102.43}} &106.20& 146.93 & 110.59 &{\textbf{81.141}}   \\
		Entropy↑ & 6.7025 & {\textbf{7.0488}} &6.9412 &6.3769  &{\underline{6.9907}}  \\  
		BIQME↑  & 0.4552 & 0.4722 & {\underline{0.4684}}   &0.3895   &{\textbf{0.5112}}  \\
		\hline
		\multicolumn{6}{c}{(b) Quantitative comparison }\\
	\end{tabular}
        \vspace{-0.2cm}
	\caption{\textbf{Performance evaluation on \textbf{\textit{nighttime haze}} scenes}. }
	\label{fig:Nighttime}
        \vspace{-0.2cm}
\end{figure}

\section{Exploring Algorithmic Property}\label{sec:property}
CoA presents an effective learning strategy to improve efficiency and adaptability for real-world scenes, without assumptions on the synthetic domain or teacher model. We analyze CoA's properties, highlighting its stability across synthetic domains and flexibility with different models.
\subsection{Stability across Various Synthetic Domains}
We perform three synthetic benchmarks (RESIDE~\cite{li2018benchmarking} (the outdoor subset), Haze4K~\cite{liu2021synthetic}, and THaze~\cite{feng2024bridging}) and three real-world datasets (RTTS~\cite{li2018benchmarking}, URHI~\cite{li2018benchmarking}, and FATTAL~\cite{fattal2014dehazing}) to assess our stability. As shown in the left section of Table~\ref{tab:PropertyQuan}, CoA consistently improves scores across four no-reference image quality metrics. Notably, due to the diverse scenes in THaze, CoA achieves near-optimal performance within the synthetic domain. We use THaze as the synthetic domain in our comparative experiments.
Fig.~\ref{fig:PropertyQual} (left) compares visual results.  The dehazing effects of CoA is significant, especially in distant regions highlighted by arrows. Overall, CoA improves the original teacher model across synthetic domains, demonstrating its stability.

\subsection{Flexibility with Different Dehazing Models}
CoA's teacher model-irrelevant nature allows us to enhance the adaptability of existing dehazing models across various scenes. We applied CoA to three dehazing models (MSBDN~\cite{dong2020multi}, DehazeFormer~\cite{song2023vision}, and DEA~\cite{chen2024dea}) as teacher models. The right section of Table~\ref{tab:PropertyQuan} shows significant performance improvements before and after applying CoA across different models and test settings. Fig.~\ref{fig:PropertyQual} (right) compares visual results using DehazeFormer and DEA, highlighting areas where CoA enhances dehazing for more visually appealing outcomes. In summary, CoA demonstrates notable flexibility with various dehazing models. 
The introduction of CoA significantly reduces the model's parameter count, improving computational efficiency. Specifically, the parameters in the three dehazing models decrease by 94.06\%, 86.74\%, and 73.89\%, respectively. This reduction accelerates inference speed and lowers resource requirements, making the models more lightweight and better suited for practical applications.

\begin{figure}[t]
	\centering
	\footnotesize
	\begin{tabular}{c} 
		\includegraphics[width=0.455\textwidth]{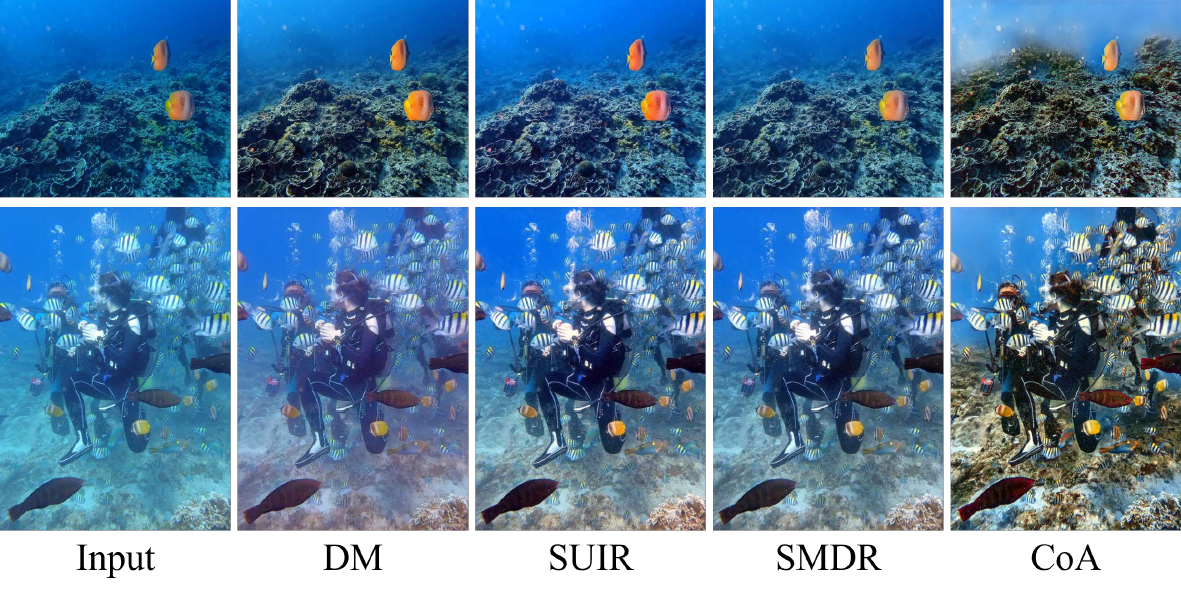}\\
		(a) Qualitative comparison 
	\end{tabular}
	\renewcommand\arraystretch{1.1} 
	\setlength{\tabcolsep}{2.5mm}
	\footnotesize
	\centering
	\begin{tabular}{|l|c|c|c|c|}
		\hline
		\multirow{2}{*}{Metrics}&\multirow{2}{*}{\makecell{DM\\\scriptsize{(\textit{ACM MM'23})}}}&\multirow{2}{*}{\makecell{SUIR\\\scriptsize{(\textit{CVPR'23})}}}&\multirow{2}{*}{\makecell{SMDR\\\scriptsize{(\textit{AAAI'24})}}}&\multirow{2}{*}{CoA}\\
		&&&&\\
		\hline
		UIQM$\uparrow$&\underline{4.0965}&3.9853&4.0314&\textbf{4.1284}\\
		CCF$\uparrow$&26.651&\textbf{31.417}&28.439&\underline{30.878}\\
		\hline
		\multicolumn{5}{c}{(b) Quantitative comparison }\\
	\end{tabular}
        \vspace{-0.2cm}
	\caption{\textbf{Performance evaluation on \textbf{\textit{underwater}} scenes}. }
	\label{fig:Underwater}
        \vspace{-0.3cm}
\end{figure}

\section{Experimental Results}

\begin{table*}[t]
	\renewcommand\arraystretch{1.3} 
	\setlength{\tabcolsep}{2.5mm}
	\centering
	\footnotesize
	\begin{threeparttable}[b]
	\begin{tabular}{|l |c|cccccccc|c|}
		\hline
		\multicolumn{2}{|c|}{\multirow{2}{*}{Metrics}} & \multirow{2}{*}{\makecell{SGID\\\scriptsize{(\textit{TIP'22})}}} & \multirow{2}{*}{\makecell{Dehamer\\\scriptsize{(\textit{CVPR'22})}}} & \multirow{2}{*}{\makecell{C2P\\\scriptsize{(\textit{CVPR'23})}}} & \multirow{2}{*}{\makecell{RIDCP\tnote{$\dagger$}\\\scriptsize{(\textit{CVPR'23})}}} & \multirow{2}{*}{\makecell{D4\\\scriptsize{(\textit{CVPR'22})}}} & \multirow{2}{*}{\makecell{D4+\\\scriptsize{(\textit{IJCV'24})}}} & \multirow{2}{*}{\makecell{KANet\\\scriptsize{(\textit{TPAMI'24})}}} & \multirow{2}{*}{\makecell{DEA\\\scriptsize{(\textit{TIP'24})}}} &
		\multirow{2}{*}{Ours} \\
		\multicolumn{2}{|l|}{~} &  & & & & & & & & \\
		\hline 
		\multicolumn{2}{|l|}{SIZE (M)} &13.87&132.40 &7.17 & 28.72& 10.70&10.70 &55.25 & 3.65& \textbf{1.69}\\
		\hline
		\multicolumn{2}{|l|}{FLOPs (G)} &108.40 & 48.91& 352.90& 144.43& 2.82&2.82 &4.42 &32.20 & \textbf{2.67}\\
		\hline
		\multirow{3}{*}{\rotatebox{90}{Time (ms)\;}} & 1280$\times$720 & 240.01&136.36&1169.18&630.64& 51.25& 51.25&22.93 &86.33&\textbf{24.33}\\
		~ & 1920$\times$1080  & 878.94&295.98&2531.43 &1588.26 &105.31 &105.31 & 205.40&190.78 &\textbf{52.52}\\
		~ &2560$\times$1440 &1514.82 &519.52 &4409.12 & ---&185.55 &185.55&  10631&334.78 & \textbf{92.13}\\
		\hline
	\end{tabular}
	\begin{tablenotes}
		\tiny \item[$\dagger$]This work requires excessive video memory for high-resolution images, resulting in the failure to produce results.
	\end{tablenotes}
    \vspace{-0.2cm}
	\caption{\textbf{Evaluating the computational efficiency}. Note that the size of the testing image for FLOPs calculation is 224$\times$224.}\label{tab:efficiency}
        \vspace{-0.2cm}
\end{threeparttable}
\end{table*}

\subsection{Implementation Details}
\textbf{Training settings.}
The CoA model was implemented using the PyTorch framework on a single NVIDIA RTX 3090 GPU.
 The Adam optimizer was employed to update each module’s parameters, with $\beta _1$, $\beta _2$, and $\varepsilon$ set to 0.9, 0.999, and $1e^{-8}$, respectively. The initial learning rate was set to $1e^{-4}$ and gradually decayed to $1e^{-6}$ using a cosine annealing schedule.
The algorithm was trained on RGB channels with data augmentation via random 90°, 180°, and 270° rotations and horizontal flipping. We cropped random 256 × 256 sub-regions from images, further expanding the training set. Using a pre-trained Res2Net encoder, the teacher model achieved strong performance after just 20 epochs.

\noindent\textbf{Benchmarks and metrics.}
 Three synthetic datasets (Haze4K, RESIDE (the outdoor subset), and THaze served as the synthetic domain, while three real-world datasets (RTTS, Fattal, and URHI were used for testing in the real domain. 
We employed four objective metrics for real-world haze scenes: FADE~\cite{DBLP:journals/tip/ChoiYB15}, PM2.5~\cite{DBLP:journals/tie/GuQL19}, Entropy, and BIQME~\cite{DBLP:journals/tnn/GuTQL18}. For underwater enhancement, we used UIQM and CCF to assess image quality.

\noindent\textbf{Compared methods.}
We evaluated our CoA method against state-of-the-art techniques across daytime, dusty, nighttime, and underwater conditions. For daytime and dusty conditions, we compared CoA with SGID~\cite{DBLP:journals/tip/BaiPXT22}, Dehamer~\cite{DBLP:conf/cvpr/GuoYACRL22}, C2P~\cite{DBLP:conf/cvpr/ZhengZHD023}, KANet~\cite{fengtpami24}, DEA, D4~\cite{yang2022self}, D4$^+$\cite{DBLP:journals/ijcv/YangWGT24}, and RIDCP~\cite{DBLP:conf/cvpr/WuDGCL23}. Additionally, we assessed CoA against multi-weather restoration models: AirNet~\cite{DBLP:conf/cvpr/LiLHW0022}, WeatherDiff~\cite{ozdenizci2023}, and DiffUIR~\cite{zheng2024selective}. For nighttime dehazing, we benchmarked against GS~\cite{DBLP:conf/iccv/LiTB15}, MRP~\cite{DBLP:conf/cvpr/ZhangCFKC17}, OSFD~\cite{DBLP:conf/mm/ZhangCZT20}, and GAPSF~\cite{DBLP:111}, and for underwater scenes, we compared with DM~\cite{tang2023underwater}, SUIR~\cite{huang2023contrastive}, and SMDR~\cite{zhang2024synergistic}.

\begin{figure}[t]
	\centering
	\footnotesize
	\begin{tabular}{c} 
		\includegraphics[width=0.43\textwidth]{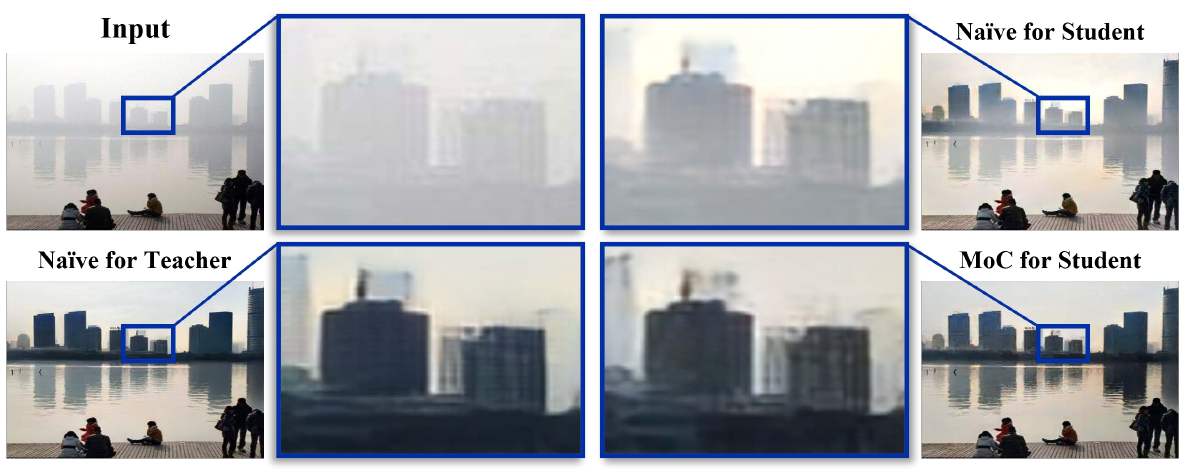}\\
		(a) Qualitative comparison 
	\end{tabular}
	\renewcommand\arraystretch{1.2} 
	\setlength{\tabcolsep}{0.8mm}
	\scriptsize
	\centering
	\begin{tabular}{|l|c|c|c|c|c|}
		\hline
		Model&FADE$\downarrow$&PM2.5$\downarrow$&BIQME$\uparrow$&SIZE (M)&FLOPs (G)\\
		\hline
		Naive for Teacher &1.21&131.77&\textbf{0.58}&52.83&18.56\\
		\hline
		Naive for Student &1.23&135.76&0.57&\textbf{1.69}&\textbf{2.67}\\
		\hline
		MoC for Student &\textbf{1.13}&\textbf{124.43}&\textbf{0.58}&\textbf{1.69}&\textbf{2.67}\\
		\hline
		\multicolumn{6}{c}{\footnotesize (b) Quantitative comparison }\\
	\end{tabular}
        \vspace{-0.2cm}
	\caption{\textbf{Effects of MoC phase}.}
	\label{fig:MoC}
        \vspace{-0.3cm}
\end{figure}

\subsection{Evaluation on Regular Scenes}

\noindent\textbf{Quantitative comparisons.}
In Fig.~\ref{fig:Quantitative}, we presented a comparison of eight state-of-the-art dehazing algorithms with our proposed method, using four authoritative unsupervised evaluation metrics. It is evident that our method ranked among the top in almost all metrics, achieving optimal or near-optimal performance in most cases. 

\noindent\textbf{Qualitative comparisons.}
As shown in Fig.~\ref{fig:Daytime}, the supervised methods (SGID, C2P, Dehamer, DEA) and unsupervised methods (D4, D4$^+$) struggle with haze removal and generalization. RIDCP causes color distortion and over-dehazing, especially in sandstorm and colorful haze scenes. KANet has limitations in detail and color restoration. In contrast, CoA excels at haze removal while preserving fine textures, resulting in more natural and realistic images.

\noindent\textbf{Comparisons with multi-weather restorers.}
Here we compared CoA with state-of-the-art multi-weather degradation methods. As shown in Table~\ref{tab:MultiTask}, CoA ranks second in one metric, outperforming others in all others, highlighting its superiority. Fig.~\ref{fig:MultiTask} reveals that while existing methods excel in specific scenarios, they show limitations in handling the full spectrum of haze conditions.

\noindent\textbf{Efficiency comparisons.}
Table~\ref{tab:efficiency} presents a comprehensive comparison of our proposed CoA method with existing approaches, evaluating parameters, floating-point operations per second (FLOPs), and processing time across various image resolutions. The results clearly show that our method has the smallest number of parameters and FLOPs among all the evaluated methods. Notably, as the image resolution increases, our method demonstrates a significant advantage in processing time, indicating that our model achieves a favorable balance between performance and complexity.

\begin{figure}[t]
	\centering
	\footnotesize
	\begin{tabular}{c} 
		\includegraphics[width=0.43\textwidth]{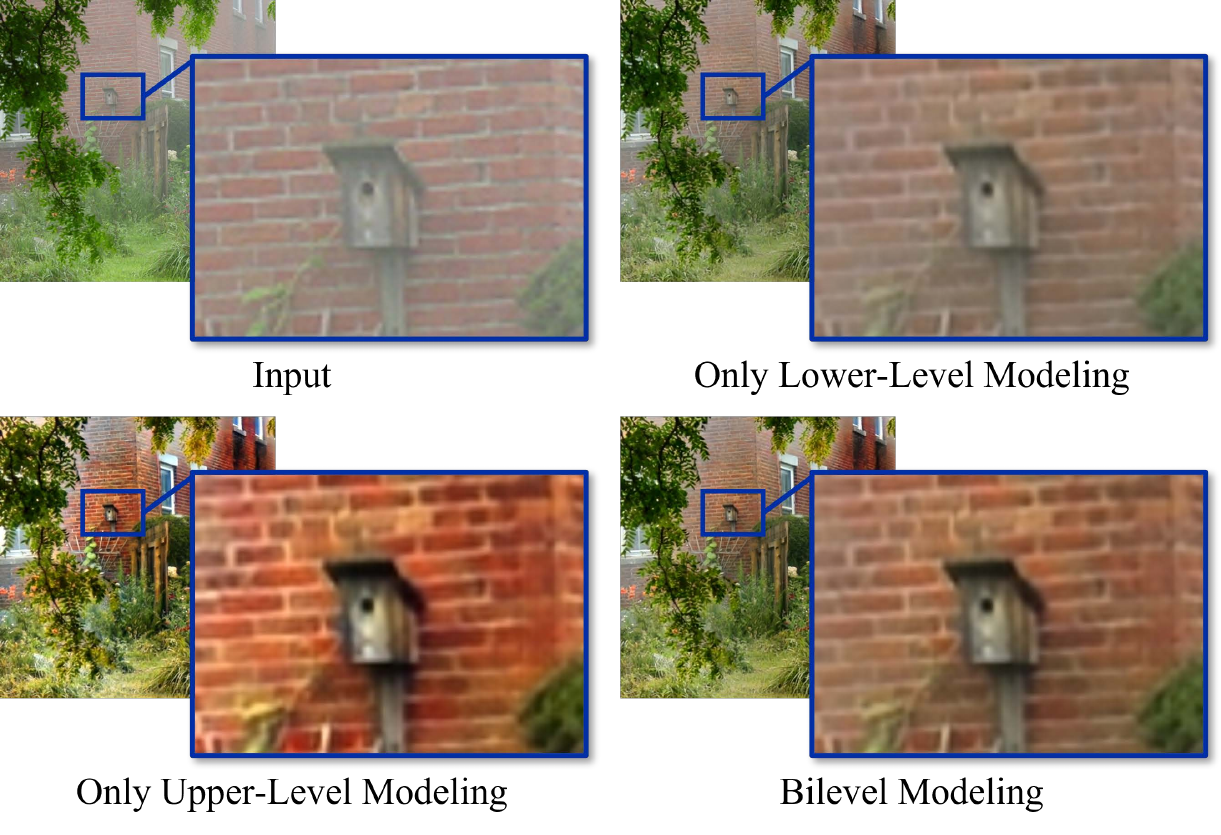}\\
	\end{tabular}
	\vspace{-0.2cm}
	\caption{\textbf{Necessity of bilevel modeling}. }
	\label{fig:Necessity}
	\vspace{-0.3cm}
\end{figure}

\subsection{Adaptability Verification}

\noindent\textbf{Nighttime haze scene.}
Compared to state-of-the-art nighttime dehazing algorithms on the NHRW dataset (Fig.~\ref{fig:Nighttime} (a)), many existing methods fail to handle overexposure or low-light conditions, resulting in overly dark or distorted images due to noise and artificial light scattering. In contrast, CoA excels in real-world nighttime haze scenes, effectively removing haze, suppressing noise, and preserving details. As shown in Fig.~\ref{fig:Nighttime} (b), our method outperforms others across three authoritative objective metrics

\noindent\textbf{Underwater scene.}
Compared to state-of-the-art underwater dehazing algorithms, DM results in reddish and blurry images, failing to correct color distortion effectively, while SUIR and SMDR struggle to fully eliminate haze. As shown in Fig.~\ref{fig:Underwater}, our method significantly reduces haze impact and enhances clarity and detail visibility, closely aligning with human visual perception standards.

\begin{figure}[t]
	\centering
	\footnotesize
	\begin{tabular}{c} 
		\includegraphics[width=0.455\textwidth]{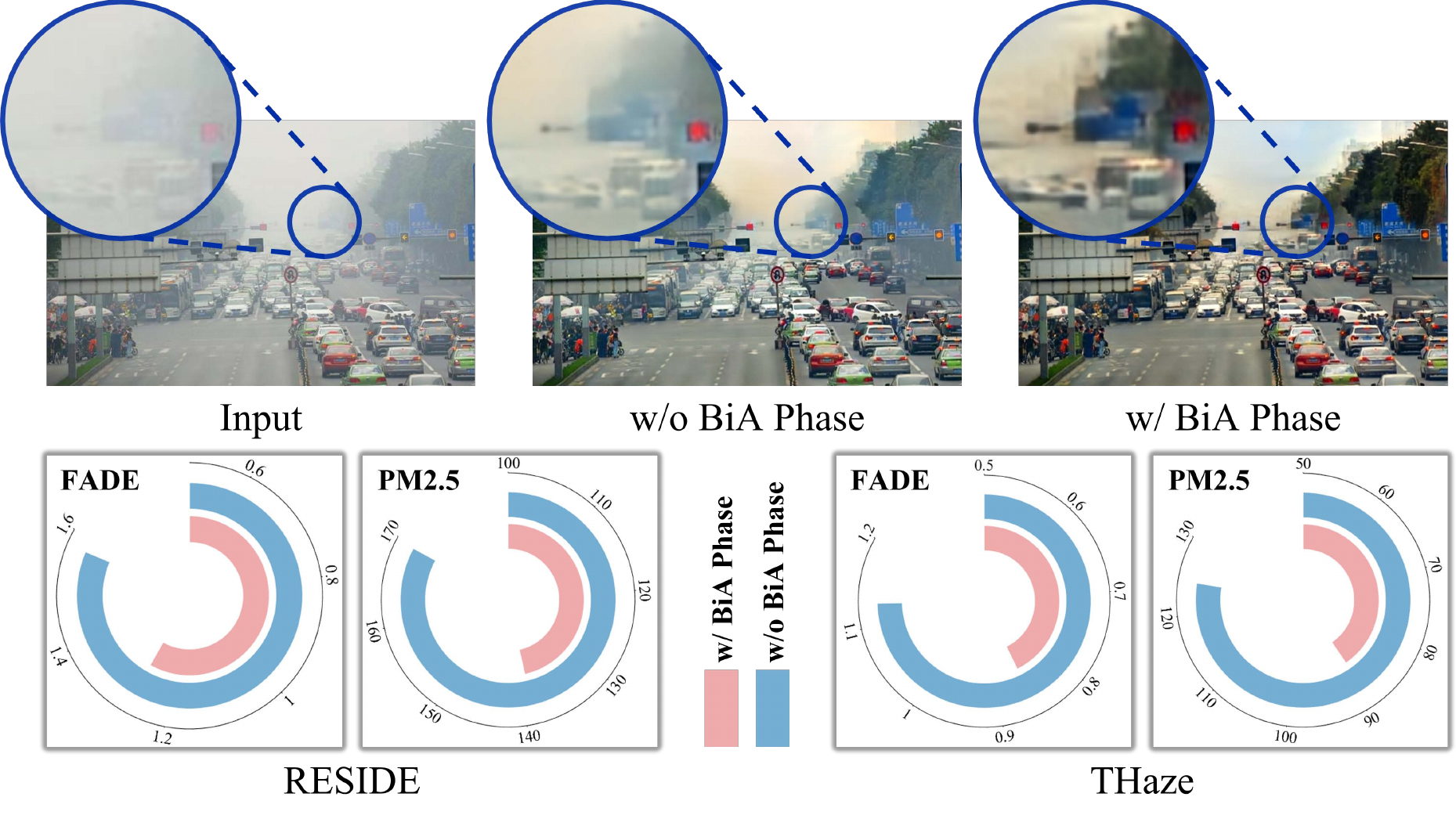}\\
	\end{tabular}
        \vspace{-0.2cm}
	\caption{\textbf{Effects of BiA phase}. }
	\label{fig:BiA}
        \vspace{-0.2cm}
\end{figure}

\section{Analytical Experiments}

\subsection{Effects of MoC Phase}
Without feature transfer from the ``naive for teacher" (\textit{naive: end-to-end training, the same below}) model via compression, the ``naive for student" model struggles to capture core data patterns, particularly in complex or extreme weather scenarios, resulting in limited dehazing. As illustrated in Fig.~\ref{fig:MoC}, while ``MoC for student" may not restore details as effectively as ``naive for teacher", it successfully inherits high-quality features. Even under resource constraints, it surpasses ``naive for student" in dehazing and performs well on unsupervised metrics.
\subsection{Necessity of Bilevel Modeling}\label{sec:necessity}
Here we highlight the critical role of bilevel modeling by training the same network model on the THaze dataset. As shown in Fig.~\ref{fig:Necessity}, the results on the real test set show marked contrasts: employing only the lower-level modeling strategy results in suboptimal dehazing, while relying solely on the upper-level strategy with CLIP fine-tuning produces unnatural effects with notable color distortion. Different from them, the bilevel modeling strategy enables both effective dehazing and natural visual quality, achieving a balance between dehazing performance and perceptual realism.

\begin{figure}[t]
	\centering
	\footnotesize
	\begin{tabular}{c} 
		\includegraphics[width=0.455\textwidth]{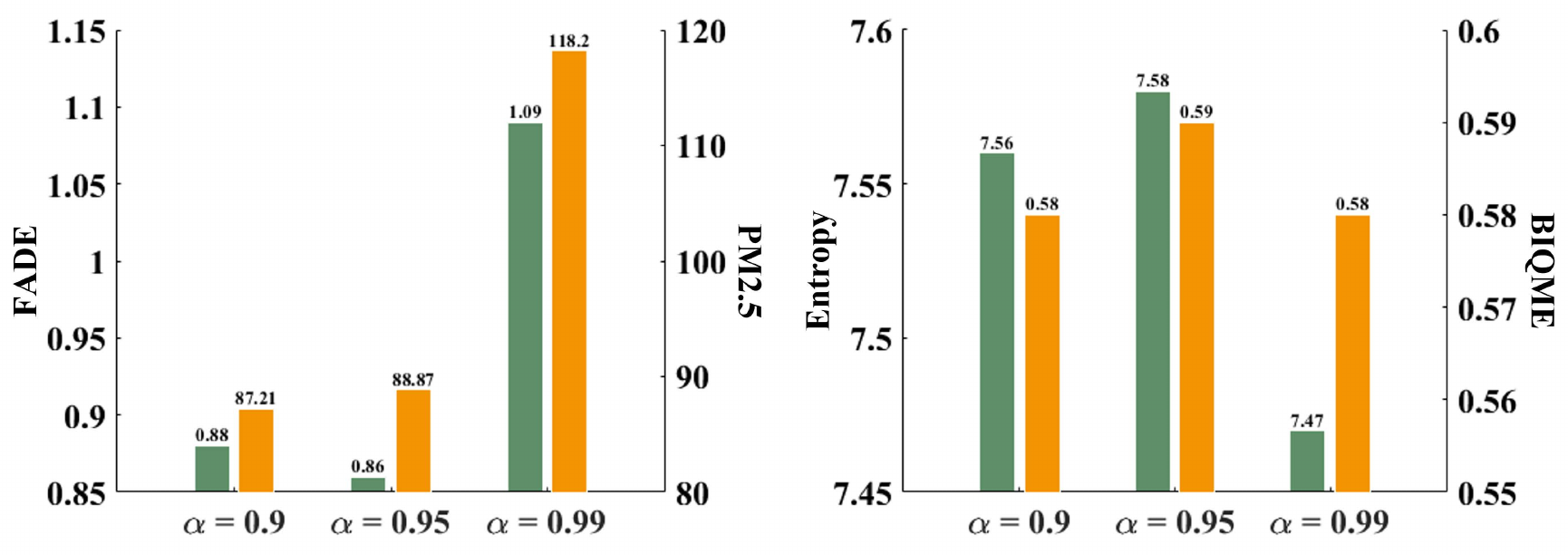}\\
	\end{tabular}
        \vspace{-0.2cm}
	\caption{\textbf{Parameters Analysis}. The RTTS dataset is adopted.}
	\label{fig:ParametersAnalyses}
        \vspace{-0.2cm}
\end{figure}

\subsection{Effects of BiA Phase}
In Fig.~\ref{fig:BiA}, we evaluate effects of the BiA phase. The results clearly show significant improvements in both FADE and PM2.5 metrics following the BiA phase. Notably, in images with complex backgrounds or high noise levels, our model outperforms the baseline, demonstrating the BiA phase's effectiveness in capturing key features and improving robustness in challenging scenarios.

\subsection{Parameters Analyses}
We test different values of the parameter $\alpha$ during the BiA phase, as shown in Fig.~\ref{fig:ParametersAnalyses}. The results indicate that $\alpha = 0.95$ yields almost optimal performance. Notably, this setting achieves the fastest convergence and exhibits superior stability and accuracy across multiple evaluation metrics, clearly outperforming other parameter settings.

\begin{figure}[t]
	\centering
	\footnotesize
	\begin{tabular}{c} 
		\includegraphics[width=0.455\textwidth]{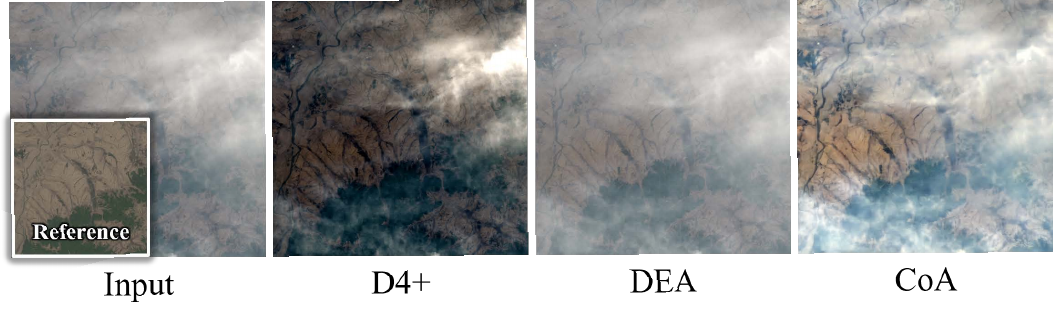}\\
	\end{tabular}
	\vspace{-0.2cm}
	\caption{\textbf{Limitations}. The example is from RS-Haze~\cite{song2023vision}. }
	\label{fig:Limitations}
	\vspace{-0.2cm}
\end{figure}

\subsection{Limitations}
While effective in many scenarios, our method struggles with remote sensing haze images due to additional imaging constraints.
As depicted in Fig.~\ref{fig:Limitations}, although it improves dehazing (outperforming DEA), non-uniform fog occlusion remains. This stems from varying haze effects across spectral bands in hyperspectral images, requiring multi-band integration for effective dehazing. 

\section{Conclusion}
This work proposes a novel compression-and-adaptation scheme to tackle the problem of unbalance between dehazing quality and computational efficiency in real image dehazing. Following the divide-and-conquer paradigm, we design a new CoA learning strategy that first compresses the predefined model and then adapts it to unlabeled, diverse real-world scenes.
We not only make a thorough exploration to take on the excellent properties of CoA, but also we perform extensive experiments to indicate our CoA method's superiority in terms of image dehazing quality and computational efficiency. 

\section*{Acknowledgments}
This work was supported by the Postdoctoral Fellowship Program of CPSF (No. GZB20240098), China Postdoctoral Science Foundation (No. 2023M740491), Dalian Science and Technology Innovation Fund-Young Tech Star (No. 2023RQ017), Fundamental Research Funds for the Central Universities (No. DUT24BS011).